\documentclass[runningheads]{llncs}

\usepackage{eccv}

\usepackage{eccvabbrv}

\usepackage{graphicx}
\usepackage{booktabs}
\usepackage{multirow}
\usepackage{wrapfig}
\usepackage{float}
\usepackage{tabularx}
\usepackage{pifont} 
\newcommand{\xmark}{\ding{55}}
\usepackage[accsupp]{axessibility}  

\usepackage{hyperref}

\usepackage{orcidlink}

\begin{document}

\title{CanvasMAR: Improving Masked Autoregressive Video Prediction With Canvas} 

\titlerunning{CanvasMAR}






\author{Zian Li\inst{1,2} \and Muhan Zhang\inst{1,3}\thanks{Corresponding Author.}}

\authorrunning{Z. Li and M. Zhang}

\institute{Institute for Artificial Intelligence, Peking University, Beijing, China \and
School of Intelligence Science and Technology, Peking University, Beijing, China \and
State Key Laboratory of General Artificial Intelligence, Peking University, Beijing, China\\
\email{zian@stu.pku.edu.cn,} \email{muhan@pku.edu.cn}
}

\maketitle

\begin{abstract}
Masked autoregressive models (MAR) have emerged as a powerful paradigm for image and video generation, combining the flexibility of masked modeling with the expressiveness of continuous tokenizers. However, when sampling individual frames, video MAR models often produce highly distorted outputs due to the lack of a structured global prior, especially when using only a few sampling steps. To address this, we propose \textbf{CanvasMAR}, a novel autoregressive video prediction model that predicts high-fidelity frames with few sampling steps by introducing a canvas—a blurred, global one-step prediction of the next frame that serves as a non-uniform mask during masked generation. The canvas supplies global structure early in sampling, enabling faster and more coherent frame synthesis. To further stabilize autoregressive sampling, we propose an easy-to-hard curriculum via a motion-aware sampling order that synthesizes relatively stationary regions before attending to highly dynamic ones. We also integrate compositional classifier-free guidance that jointly strengthens the canvas and temporal conditioning to improve generation fidelity. Experiments on the BAIR, UCF-101, and Kinetics-600 benchmarks demonstrate that CanvasMAR produces higher-quality videos with fewer autoregressive steps. On the challenging Kinetics-600 dataset, CanvasMAR achieves remarkable performance among autoregressive models and rivals advanced diffusion-based methods.
\end{abstract}

\section{Introduction}
\label{sec:intro}

Masked generative models have achieved remarkable success in both image generation~\cite{chang2022maskgit, li2023mage, li2024autoregressive} and video generation~\cite{yu2023magvit, zhou2025taming, deng2024autoregressive}. By representing images as tokens and generating them set by set in a random order,\footnote{For models such as MaskGiT~\cite{chang2022maskgit}, the generation order is not strictly random because it also relies on token probabilities interpreted as confidence scores. However, we adopt a generalized notion of ``random'' to contrast with strictly predefined orders like raster scan.} these models achieve higher fidelity and scalability than approaches that rely on predefined orders~\cite{Fan2024FluidSA}. Among these approaches, Masked Autoregressive Models (MAR)~\cite{li2024autoregressive, Fan2024FluidSA}\footnote{We adopt the generalized notion of ``autoregressive'' introduced in~\cite{li2024autoregressive}, which treats masked generation as autoregression along the spatial axis in random order.} pioneer the generation of continuous image tokens with a diffusion head, in contrast to prior works that use discrete tokens~\cite{chang2022maskgit, yu2023magvit}. This eliminates the quantization errors inherent in discrete-token masked generative models and avoids the challenges of training high-quality VQ-VAEs~\cite{van2017neural, esser2021taming, yu2023language, han2025infinity}, demonstrating strong potential both theoretically and empirically.

\begin{figure*}[t]
\centering
\includegraphics[width=0.9\linewidth]{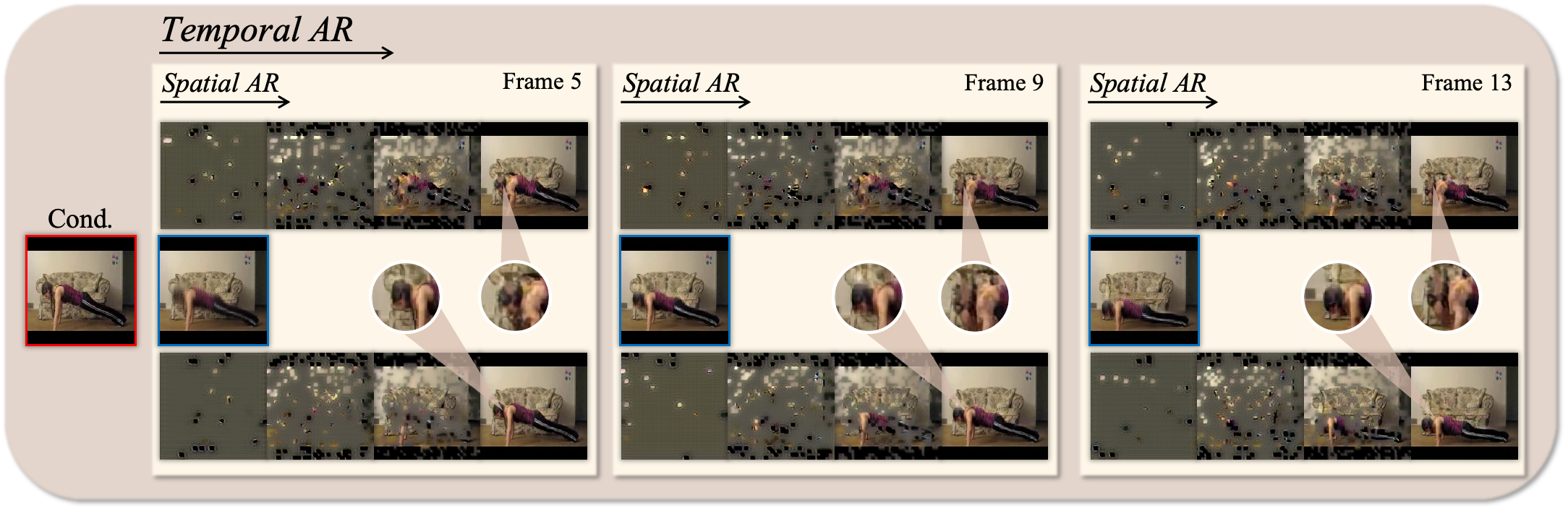}
\caption{Video generation with Masked Autoregressive Models using only 8 autoregressive steps per $32 \times 32$ latent frame. Frames with red borders (left) denote the conditioning inputs. Without a canvas (top row), the model quickly loses global coherence for the body. Our approach (second and bottom rows) mitigates this by first predicting a blurred canvas (shown in the second row with blue borders) that captures coarse motion and global structure. The canvas then replaces the uniform mask in the masked autoregressive model, allowing global fidelity to be preserved even under an aggressive sampling regime.}
\label{fig:canvas}
\end{figure*}

Extending the MAR paradigm—continuous image tokens with random-order autoregressive generation—to video generation is promising, as evidenced by several recent studies~\cite{zhou2025taming, deng2024autoregressive, yu2025videomar}. However, a key challenge remains: MAR follows the MaskGIT~\cite{chang2022maskgit} sampling paradigm, which begins from a fully masked image and autoregressively generates tokens in a random order. In early stages, when only a few tokens have been generated, the model must generate small sets of tokens at each step to maintain quality. The set size gradually increases only as more tokens are revealed, typically following a cosine schedule~\cite{chang2022maskgit}. This creates a \emph{trade-off between fidelity and sampling speed}: achieving high-quality generation typically requires significantly many sampling steps, while using only few steps often results in highly distorted outputs—especially for videos, where the temporal dimension further amplifies the issue. As illustrated in \cref{fig:canvas}, a simple video MAR model exhibits noticeable degradation after only 8 spatial autoregressive steps when generating a $32 \times 32$ latent frame ($256 \times 256$ pixel frame), and the problem worsens as more frames are rolled out.

In this paper, we introduce \textbf{CanvasMAR}, a masked autoregressive video prediction model that sustains \emph{high fidelity} while using \emph{few autoregressive steps} per frame. CanvasMAR operates through a two-level autoregressive process: temporally, it generates frames sequentially; spatially, it partitions each frame into image tokens and produces them set by set in an adaptive, motion-aware order. To maintain fidelity with few sampling steps, CanvasMAR first predicts a blurred next frame—the \textit{canvas} (see~\cref{fig:canvas})—with only one model pass, and replaces the uniform masks in conventional masked generators with these non-uniform \textit{canvas tokens}. The canvas captures the global structure of the target frame, enabling more aggressive yet coherent synthesis. To stabilize sampling, we further introduce an adaptive, motion-aware sampling order for next-frame generation: the model first synthesizes low-motion (near-stationary) regions, then progressively attends to highly dynamic areas, yielding an easy-to-hard curriculum that stabilizes autoregressive sampling and suppresses motion artifacts. We also develop compositional classifier-free guidance~\cite{ho2022classifier, du2024compositional, brooks2023instructpix2pix} that jointly strengthens the canvas and temporal conditioning, further improving fidelity.

Experiments on standard video prediction datasets—including the lab-scale BAIR~\cite{unterthiner2018towards, ebert2017self}, the moderate-scale UCF-101~\cite{soomro2012ucf101}, and the large-scale Kinetics-600~\cite{carreira2018short} benchmarks—validate the effectiveness of our approach. On BAIR and Kinetics-600, CanvasMAR delivers significant gains over prior autoregressive models such as MAGI~\cite{zhou2025taming} and rivals strong diffusion-based baselines like DFoT~\cite{song2025history}, underscoring the potential of global structure-aware masked generative models.

\section{Related Work}
\label{sec:related}

\subsection{Diffusion-Based Video Generative Models} Diffusion models have demonstrated remarkable success in image generation~\cite{ho2020denoising, rombach2022high, song2020score}, and have since been extended to video generation by treating the temporal axis as an additional dimension alongside the spatial ones~\cite{ho2022video, brooks2024video, kong2024hunyuanvideo, wan2025wan, yang2024cogvideox, ma2024latte}. While effective, this paradigm generates entire videos as a whole, which leads to high latency and limited interactivity—two critical drawbacks for applications such as interactive simulators and game environments~\cite{feng2024matrix, bruce2024genie}. To mitigate these issues, recent research has adopted an autoregressive approach, wherein videos are generated sequentially by conditioning on previously generated clean frames. This design enables interactivity and supports the generation of longer, temporally coherent videos~\cite{ruhe2024rollingdiffusionmodels, song2025history, yin2025slow, gupta2024photorealistic, zhang2025generative, liu2025worldweaver, liu2025rolling, huang2025self}. However, these methods still remain diffusion-based, which generate individual frames or chunks as a whole through iterative frame-level denoising. In contrast, token-based autoregressive models that this work focuses on directly learn token-level distributions and produce image tokens, making them more natively compatible with large language models and thus better suited for multimodal model designs~\cite{wang2024emu3,jin2024video}.

\subsection{Autoregressive Video Generation} An alternative line of work decomposes video frames into image tokens, shifting the generative task from producing full frames to predicting sequences of tokens. Many approaches in this direction adopt quantized tokens, typically following paradigms such as next-token prediction~\cite{yan2021videogpt, hong2022cogvideo, sun2024autoregressive, wang2024loong, wang2024emu3, jin2024video} or masked generative modeling~\cite{gupta2022maskvit, yu2023magvit, yu2023language, bruce2024genie, wang2024worlddreamer, villegas2022phenaki}. However, as observed by Yu \etal.~\cite{yu2023language}, these paradigms rely on carefully designed and trained VQ-VAEs~\cite{van2017neural, esser2021taming}, which inevitably introduce quantization errors. Inspired by Masked Autoregressive Models (MAR)~\cite{li2024autoregressive, Fan2024FluidSA}, a growing body of work—representative works including NOVA~\cite{deng2024autoregressive}, VideoMAR~\cite{yu2025videomar}, and MAGI~\cite{zhou2025taming}—adapts MAR to video generation, which directly generate continuous image tokens and achieve remarkable generation performance.

\subsection{Cascaded Generation} The closest line of work that shares a similar idea with \textit{canvas} is cascaded generation. Cascaded generation frameworks decompose synthesis into stages that progressively increase resolution or temporal span~\cite{saharia2022image, gu2023matryoshka, wang2025lavie, ho2022cascaded, zhang2025flashvideo, zheng2025himar, he2024venhancer}. Early image pipelines such as Cascaded Diffusion~\cite{ho2022cascaded} train a low-resolution base model to capture global semantics before invoking specialized super-resolution and detail-enhancement modules. Such a coarse-to-fine strategy improves sample quality but requires separate networks, schedulers, and guidance hyperparameters per stage. The same underlying insight also extends to autoregressive approaches like VAR~\cite{tian2024visual} for next-scale prediction and to masked generative models like HiMAR~\cite{zheng2025himar}, though implementations differ. Recent large-scale video systems includes FlashVideo~\cite{zhang2025flashvideo} and Waver~\cite{zhang2025waver} also adopt similar ideas for fast and high-fidelity video generation. However, \textbf{CanvasMAR is fundamentally different from these cascaded methods}: these works produce \textit{multi-step generated samples} $x\sim p(x)$ for each low-resolution image/frame $x$, whereas CanvasMAR produces a \textit{one-step predicted expectation} $\mathbb{E}[x]$ for the next frame $x$, which is computationally much more efficient and tailored for video MAR models.

\section{Preliminaries}
\label{sec:prelim}

\subsection{Factorizing Video Generation}

In this paper, we focus on autoregressive video prediction. 
A video is denoted as $V \in \mathbb{R}^{N \times H \times W \times 3}$, which consists of $N$ frames $[f^{(1)}, f^{(2)}, \ldots, f^{(N)}]$ of height $H$ and width $W$. We employ latent generation~\cite{rombach2022high}; however, for simplicity, we retain pixel-level notation.  
Autoregressive video generation models the video distribution via the following \textit{temporal} factorization:
\begin{equation}
p(V) =  \prod_{i=1}^{N} p\big(f^{(i)} \mid f^{(<i)}\big),
\end{equation}
where $f^{(<i)} = \{f^{(1)}, f^{(2)}, \ldots, f^{(i-1)}\}$ denotes all preceding frames.  
Each frame $f^{(i)}$ is further decomposed into a sequence of image tokens in raster order, $f^{(i)} = [x^{(i)}_1, x^{(i)}_2, \ldots, x^{(i)}_n]$.  
In standard autoregressive generation~\cite{yan2021videogpt, hong2022cogvideo}, tokens are generated in raster order token-by-token:
\begin{equation}
p(f^{(i)} \mid f^{(<i)}) = \prod_{j=1}^{n} p\big(x^{(i)}_j \mid f^{(<i)}, x^{(i)}_{<j}\big),
\end{equation}
where $x^{(i)}_{<j} = \{x^{(i)}_1, x^{(i)}_2, \ldots, x^{(i)}_{j-1}\}$ denotes the tokens already generated within frame $i$. 

Masked generative paradigms~\cite{chang2022maskgit, li2024autoregressive, deng2024autoregressive} generate tokens set by set in a random order.  
Specifically, the $n$ tokens are randomly partitioned into $K$ sets of sizes $[n_1, n_2, \ldots, n_K]$, with cumulative index $s_k = \sum_{m=1}^k n_m$ and $s_0 = 0$, $s_K = n$.  
This factorization introduces \textit{spatial} autoregression:
\begin{equation}
p(f^{(i)} \mid f^{(<i)}) = \prod_{k=1}^{K} p\big(X^{(i)}_k \mid f^{(<i)}, X^{(i)}_1, X^{(i)}_2, \ldots, X^{(i)}_{k-1}\big),
\label{eq:mg}
\end{equation}
where each token set is defined as $X^{(i)}_k = \{x^{(i)}_{\tau(s_{k-1}+1)}, x^{(i)}_{\tau(s_{k-1}+2)}, \ldots, x^{(i)}_{\tau(s_k)}\}$ where $\tau \sim \mathcal{S}_n$ denotes a random permutation sampled from the symmetric group $\mathcal{S}_n$, i.e., the set of all permutations of the index sequence $[1, 2, \ldots, n]$. This sampling introduces stochasticity into the generation process.

\subsection{Masked Autoregressive Models with Continuous Tokens}

Unlike conventional autoregressive generative models that produce quantized image tokens~\cite{hong2022cogvideo, yan2021videogpt} using categorical distributions, MAR~\cite{li2024autoregressive} directly generates continuous image tokens using a diffusion head, which models token distributions conditioned on token embeddings.  
Formally, the distribution in~\cref{eq:mg} is further factorized as:
\begin{equation}
\begin{aligned}
p\big(X^{(i)}_k \mid f^{(<i)}, X^{(i)}_{<k}\big)
&:= p\big(X^{(i)}_k \mid Z^{(i)}_k\big) \cdot p\big(Z^{(i)}_k \mid f^{(<i)}, X^{(i)}_{<k}\big),
\end{aligned}
\end{equation}
where $Z^{(i)}_k = \{z^{(i)}_{\tau(s_{k-1}+1)}, z^{(i)}_{\tau(s_{k-1}+2)}, \ldots, z^{(i)}_{\tau(s_k)}\}$ denotes the token embeddings, and $X^{(i)}_{<k} = \{X^{(i)}_1, X^{(i)}_2, \ldots, X^{(i)}_{k-1}\}$.  
The conditional distribution $p(X^{(i)}_k \mid Z^{(i)}_k) = \prod_{j=s_{k-1}+1}^{s_{k}} p(x^{(i)}_j \mid z^{(i)}_j)$ models per-token outputs given the embeddings, where correlations between tokens are captured when computing the embeddings.  
In practice, $p\big(Z^{(i)}_k \mid f^{(<i)}, X^{(i)}_1, X^{(i)}_2, \ldots, X^{(i)}_{k-1}\big)$ is parameterized by a Vision Transformer~\cite{dosovitskiy2020image} that learns token-wise embeddings as $\delta$ distribution, while $p(x^{(i)}_j \mid z^{(i)}_j)$ is modeled by a lightweight MLP-based diffusion head that conditions on $z^{(i)}_j$ to generate $x^{(i)}_j$~\cite{li2024autoregressive}.

\section{Methods}
\label{sec:method}

\begin{figure*}[!ht]
\centering
\includegraphics[width=0.9\linewidth]{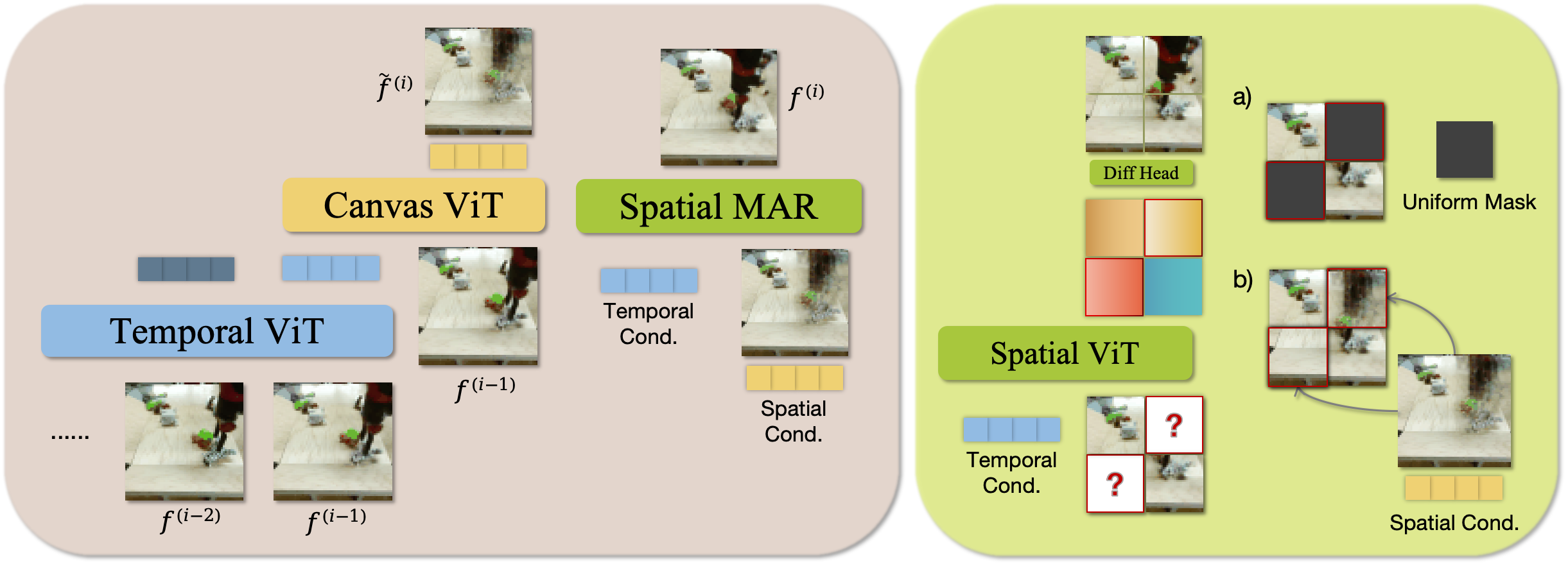}
\caption{
\textbf{Left:} Inference procedure of \textsc{CanvasMAR}. The framework consists of three modules: \emph{Temporal ViT}, \emph{Canvas ViT}, and \emph{Spatial MAR}. The Temporal ViT encodes the historical frames and produces a temporal embedding that serves as a temporal condition for both the Canvas ViT and the Spatial MAR. The Canvas ViT takes the temporal embedding together with the most recent frame to predict a coarse \emph{canvas} in one go, i.e., the initial guess of the future frame, which, along with the temporal embedding, provides spatial conditioning for the Spatial MAR. 
\textbf{Right:} Training procedure of the Spatial MAR. A Spatial ViT learns to reconstruct randomly masked regions within frames. Unlike vanilla MAR (a), which uses a single learnable mask embedding for all regions, CanvasMAR replaces this with the predicted canvas (b), offering a more informative and adaptive conditioning signal.
}
\label{fig:infer_framework}
\end{figure*}

\subsection{Factorized Temporal and Spatial Autoregression}

We design our model to generate videos through a factorized temporal-spatial autoregressive framework, as illustrated in~\cref{fig:infer_framework}. For clarity, we denote hidden representations by $z$ and refer to the temporal and canvas latents as $z_t$ and $z_c$. Specifically, CanvasMAR proceeds in two stages for generating the next frame $i$:  
1) a temporal embedding $z_t^{(i)} \in \mathbb{R}^{n \times d}$, where $n$ denotes the number of frame tokens and $d$ the hidden dimension, is autoregressively generated for frame $i$, using a Temporal ViT~\cite{dosovitskiy2020image} that conditions on all past frames $f^{(<i)}$;  
2) this temporal embedding is provided as a conditioning sequence to the Spatial MAR, implemented as a masked autoregressive model~\cite{li2024autoregressive}, which generates image tokens set by set for the next frame $f^{(i)}$.

The attention mask of the Temporal ViT follows the \emph{hybrid} structure of~Deng \etal~\cite{deng2024autoregressive}, Yu \etal~\cite{yu2025videomar}—causal across frames and bidirectional within each frame.  
This design preserves temporal causality while maintaining comprehensive spatial perception within frames, and it enables frame-level \emph{KV caching}, further accelerating generation.

We note that this temporal-spatial factorization \emph{preserves both training and inference efficiency}, unlike models that process temporal and spatial sequences uniformly~\cite{zhou2025taming,yu2025videomar}:  
1) During training, tokens within each frame can be independently masked, and all masked frames contribute to the loss in parallel, analogous to decoder-only language models~\cite{touvron2023llama, yu2023language}.  
In contrast, models without factorized autoregression can mask tokens only in the last frame of a sequence and compute loss of the only masked frame~\cite{yu2025videomar}, leading to inefficient training. 
A potential remedy, as explored in MAGI~\cite{zhou2025taming}, is to duplicate ``clean'' frames for each frame in the sequence and design a corresponding attention mask, which doubles the sequence length and substantially increases memory usage. 
2) During inference, the temporal embedding $z_t$ captures motion information necessary to guide next-frame generation.  
As a result, spatial token generation needs only attend to this fixed-length temporal embedding rather than all previously generated image tokens, whose size grows with sequence length. 

\subsection{Bridging Fast Temporal Autoregression and Slow Spatial Autoregression with Canvas}

As shown in~\cref{fig:canvas}, the absence of an initial global estimate of the target frame severely limits the size of the early token sets, i.e., $s_i$ in~\cref{eq:mg} has to be small for small $i$.  
In video generation, this further degrades both quality and speed.  
To address it, we introduce the \textit{canvas} between temporal autoregression and spatial autoregression, as illustrated in~\cref{fig:infer_framework}.

Specifically, after obtaining the temporal embedding $z_t^{(i)}$, we first pass it through the \textit{Canvas ViT} module instead of feeding it directly to the Spatial MAR. 
This module predicts a blurred, coarse version of the next frame $z_c^{(i)}$, which serves as an explicit \textit{spatial condition} for subsequent modules.  
The Spatial MAR then receives both the temporal condition $z_t^{(i)}$ and the spatial canvas embedding $z_c^{(i)}$ to predict the next frame, as shown on the right side of~\cref{fig:infer_framework}.

Concretely, the Canvas ViT takes the temporal representation $z_t^{(i)}$ together with the most recent frame $f^{(i-1)}$ as input and outputs a canvas embedding $z_c^{(i)} \in \mathbb{R}^{n \times d}$, representing an initial guess of the next frame $f^{(i)}$.  
To supervise this process, we introduce a canvas reconstruction loss:
\begin{equation}
\label{eq:canvas_loss}
\mathcal{L}_{\text{canvas}} = \mathbb{E}_{f^{(<i+1)}} \big[ \sum_{j=1}^n \| f^{\text{canvas}}_\theta (z_{c,j}^{(i)}) - x^{(i)}_j \big\|_2^2\big],
\end{equation}
where $z_{c,j}^{(i)}$ denotes the $j$-th row of $z_c^{(i)}$, and $f^{\text{canvas}}_\theta$ is a simple linear projection layer that supervises each patch embedding $z_c^{(i)}$ to match the corresponding $j$-th patch in the ground-truth future frame $x_j^{(i)}$.

We note that, unlike the Spatial MAR—which is inherently slower and stochastic—the Canvas ViT is \emph{fast and deterministic}: given the same temporal embedding $z_t^{(i)}$ and previous frame $f^{(i-1)}$, it always produces the same canvas. Theoretically, \textbf{the optimal Canvas ViT approximates the expected future frame} $\mathbb{E}_{f^{(i)}\sim p(f^{(i)} \mid f^{(<i)})}[f^{(i)}]$ given past frames, effectively producing a superposition of plausible outcomes. This initialization allows the Spatial MAR to maintain global coherence early on and quickly ``collapse'' onto one plausible future frame.

\subsection{Motion-Aware Adaptive Sampling Order}
\label{sec:motion_based_sampling}

The canvas approximates the conditional expectation of the next frame given the history. Consequently, highly dynamic regions can appear blurrier than stationary ones because they exhibit more uncertainty. Generating multiple patches in such blur regions simultaneously during early sampling steps can therefore be suboptimal, even with the canvas condition. We therefore design a motion-aware adaptive sampling order that prioritizes low-motion areas early and gradually attends to high-motion areas, implementing an easy-to-hard curriculum that stabilizes sampling and reduces motion artifacts.

Specifically, we add a lightweight \emph{staticness head} $f^{\text{static}}_\theta$ to the Canvas ViT that predicts a nonnegative scalar score per canvas patch, where larger values indicate lower motion. The score is supervised with a training-time loss that resembles confidence prediction in AlphaFold~\cite{jumper2021highly}. Formally, let the per-patch canvas reconstruction error 
\begin{equation}
\ell_j^{(i)} = \big\| f^{\text{canvas}}_\theta (z_{c,j}^{(i)}) - x^{(i)}_j \big\|_2^2.
\end{equation}
We define the target staticness as $s^{(i)}_j = \exp\big(-\ell_j^{(i)}\big)$, where $0 \leq s^{(i)}_j \leq 1$. We then train with an MSE loss
\begin{equation}
\label{eq:static_loss}
\mathcal L_{\text{static}} = \mathbb{E}_{f^{(<i+1)}} \big[\sum_{j=1}^n \big( f^{\text{static}}_\theta(z_{c,j}^{(i)}) - \text{stop}(s^{(i)}_j) \big)^2\big],
\end{equation}
where $f^{\text{static}}_\theta$ denotes a simple motion head (a linear layer followed by a sigmoid) that probes each patch embedding to predict staticness, and $\text{stop}(\cdot)$ indicates stop-gradient operation. Intuitively, with $f^{\text{canvas}}$ predicting the expectation of a future patch, $f^{\text{static}}$ estimates a quantity related to the variance of the corresponding patch.

In MAR~\cite{li2024autoregressive}, sampling proceeds in a purely random order. Although this aligns with the training distribution of masked generative modeling~\cite{li2024autoregressive}, it is suboptimal for next-frame generation because of the function-reuse approximation error (i.e., reusing a function evaluation for both regions with high and low motion)~\cite{zheng2024masked}. We therefore treat the staticness scores as confidence values that bias the sampling order toward low-motion regions in early steps and gradually revert to random order in later steps. We follow the temperature-annealing method of MaskGiT~\cite{chang2022maskgit} to retain stochasticity.

\subsection{Compositional Classifier-Free Guidance}
\label{sec:noise_aug}
Guidance is known to significantly improve generation fidelity~\cite{ho2022classifier,rombach2022high,du2024compositional,brooks2023instructpix2pix,song2025history}. In CanvasMAR, the introduction of the canvas provides a significant spatial condition in addition to the temporal one. We therefore adopt compositional classifier-free guidance to enhance robustness under this expanded conditioning setup.

For the Spatial MAR, the generative distribution is $p(f^{(i)} \mid z_c^{(i)}, z_t^{(i)})$, where $z_c^{(i)}$ denotes the canvas condition and $z_t^{(i)}$ denotes the temporal condition. We decompose the distribution as
\[
\begin{aligned}
p\big(f^{(i)} \mid z_c^{(i)}, z_t^{(i)}\big) 
\propto p\big(f^{(i)}\big)\, p\big(z_t^{(i)} \mid f^{(i)}\big)  p\big(z_c^{(i)} \mid f^{(i)}, z_t^{(i)}\big) ,
\end{aligned}
\]
and apply compositional CFG~\cite{du2024compositional, brooks2023instructpix2pix} to up-weight both the spatial and temporal posteriors:
\begin{equation}
\label{eq:comp}
\begin{aligned}
p^{(w_s, w_t)}\big(f^{(i)} \mid z_c^{(i)}, z_t^{(i)}\big)
:= p\big(f^{(i)}\big)\, p\big(z_t^{(i)} \mid f^{(i)}\big)^{w_t} p\big(z_c^{(i)} \mid f^{(i)}, z_t^{(i)}\big)^{w_s} ,
\end{aligned}
\end{equation}
where $w_s$ and $w_t$ denote the guidance scales for the spatial and temporal conditions, respectively. Intuitively, $w_s$ and $w_t$ increase the likelihood that the generated frame $f^{(i)}$ conforms to the spatial prior $z_c^{(i)}$ and the temporal-consistency constraint $z_t^{(i)}$. ~\cref{eq:comp} leads to the following score combination:
\begin{equation}
\label{eq:cfg_score_appendix}
\begin{aligned}
\mathbf{s}^{(w_s, w_t)}
&= \nabla_{f^{(i)}} \log p^{(w_s, w_t)}\big(f^{(i)} \mid z_c^{(i)}, z_t^{(i)}\big) \\
&= \mathbf{s}_{\varnothing}
  + w_t\, \big(\mathbf{s}_{t} - \mathbf{s}_{\varnothing}\big)
  + w_s\, \big(\mathbf{s}_{t,s} - \mathbf{s}_{t}\big),
\end{aligned}
\end{equation}
where $\mathbf{s}_{\varnothing} = \nabla \log p(f^{(i)})$ is the unconditional score, $\mathbf{s}_{t} = \nabla \log p(f^{(i)} \mid z_t^{(i)})$ is the temporal-only score, and $\mathbf{s}_{t,s} = \nabla \log p(f^{(i)} \mid z_t^{(i)}, z_c^{(i)})$ is the fully conditioned score.

We obtain $\{\mathbf{s}_{\varnothing}, \mathbf{s}_{t}, \mathbf{s}_{t,s}\}$ via three forward passes of the Spatial MAR head that differ only in their conditioning inputs (unconditional, temporal-only, and temporal+spatial). We then combine the predicted scores according to~\cref{eq:cfg_score_appendix} before updating tokens. 

To enable CFG at inference, we adopt classifier-free training by randomly dropping conditions: with 5\% probability we replace the spatial condition with a uniform mask~\cite{chang2022maskgit}; with 5\% probability we replace the temporal sequence with a learnable temporal vector; with 5\% probability we drop both; otherwise we keep both.

\section{Experiments}
\label{sec:experiments}

\begin{figure*}[t]
\centering
\begin{minipage}[c]{0.32\linewidth}
    \centering
    \includegraphics[width=\linewidth]{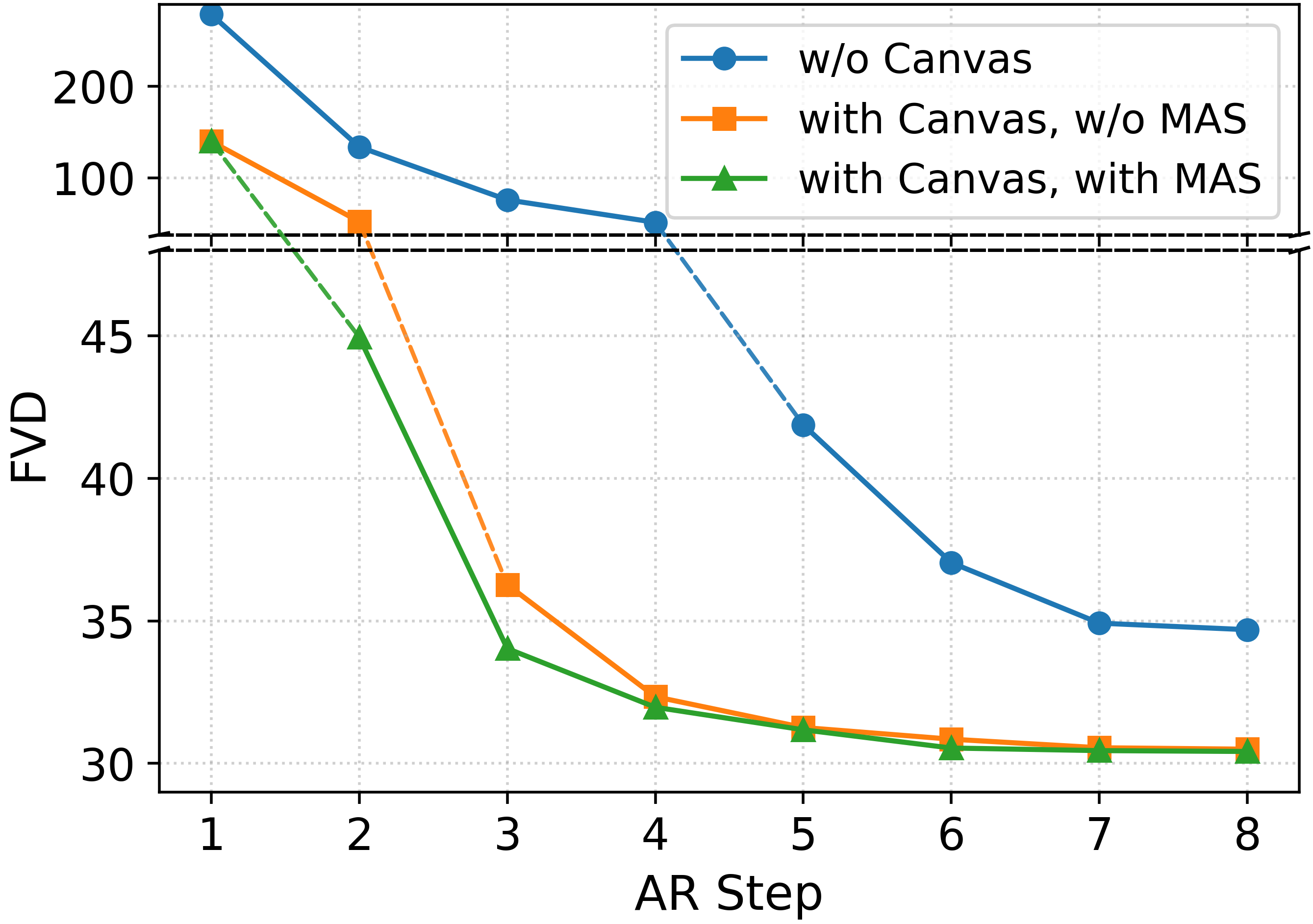}\\[0.25em]
    {\small (a)}
\end{minipage}
\hfill
\begin{minipage}[c]{0.32\linewidth}
    \centering
    \includegraphics[width=\linewidth]{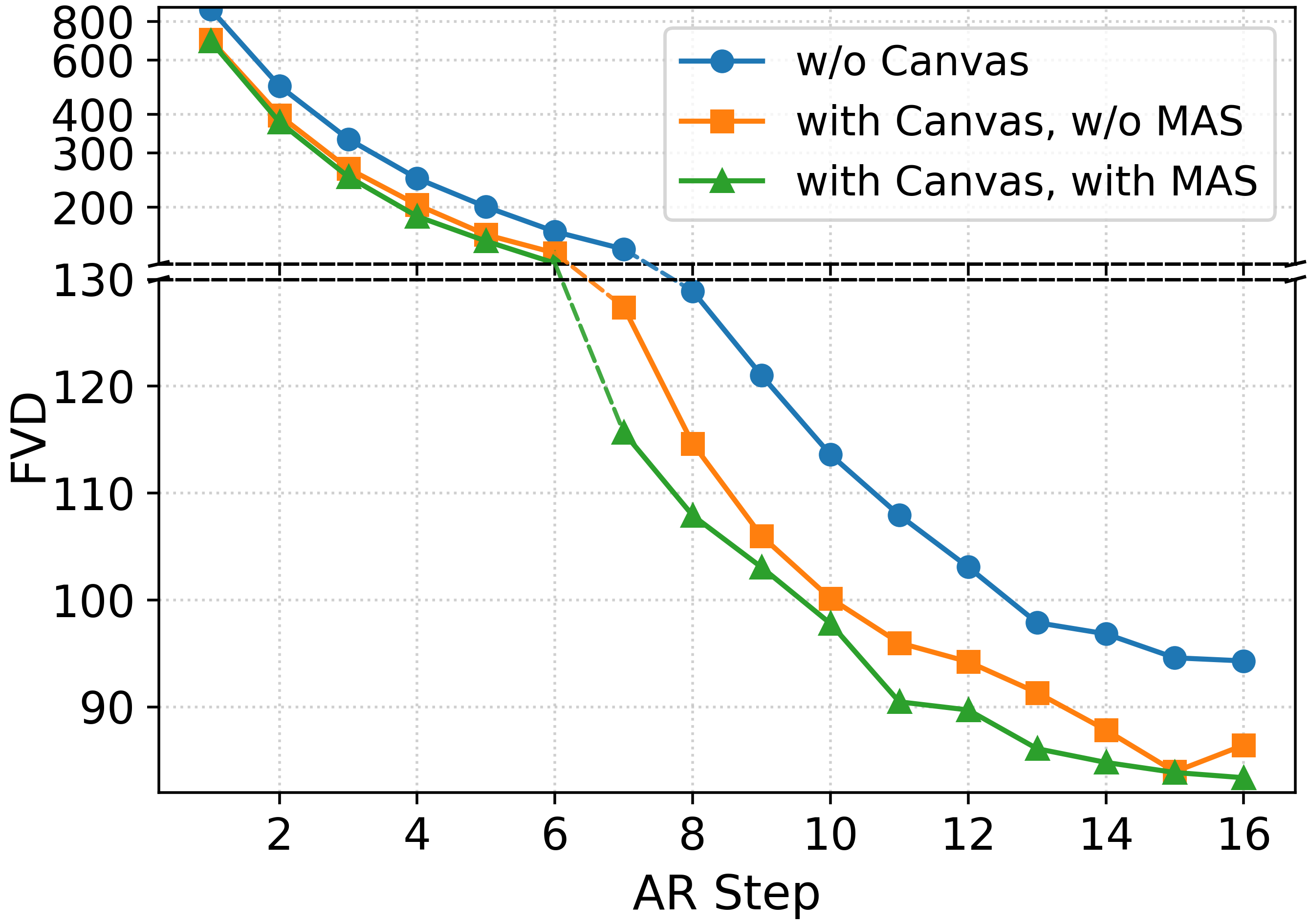}\\[0.25em]
    {\small (b)}
\end{minipage}
\hfill
\begin{minipage}[c]{0.32\linewidth}
    \centering
    \resizebox{0.9\linewidth}{!}{%
    \begin{tabular}{cccc}
    \toprule
    \textbf{sCFG} & \textbf{tCFG} & \textbf{BAIR} & \textbf{UCF101} \\
    \midrule
    \checkmark & \checkmark & \textbf{29.58} & \textbf{86.03}  \\
    \xmark     & \checkmark & 30.08 & 92.60  \\
    \checkmark & \xmark     & 30.01 & 100.90  \\
    \xmark     & \xmark     & 31.54 & 114.07  \\
    \bottomrule
    \end{tabular}%
    }\\[0.25em]
    {\small (c)}
\end{minipage}

\caption{Component-level analysis. In the first two figures, classifier-free guidance is not used, and MAS represents the motion-aware sampling order. (a) Effectiveness of canvas and motion-aware sampling on BAIR. (b) Effectiveness of canvas and motion-aware sampling on UCF-101.  (c) Effectiveness of compositional classifier-free guidance, where sCFG represents spatial (canvas) CFG, and tCFG represents the temporal CFG. The metric is FVD; the lower, the better.}
\label{fig:componet_level_analysis}
\end{figure*}

\subsection{Settings}
\subsubsection{Model Implementation.} We implement CanvasMAR on the NOVA~\cite{deng2024autoregressive} codebase with three additional changes: (1) a flow head trained with flow matching~\cite{lipman2022flow, liu2022flow} instead of DDPM~\cite{ho2020denoising}; (2) expert LayerNorm~\cite{yang2024cogvideox} that treats the conditioning sequence, clean tokens, and mask tokens separately, or MM-DiT~\cite{esser2024scaling} that treats the clean and mask tokens separately; and (3) removal of the Scale-and-Shift layer for the temporal embedding~\cite{deng2024autoregressive}.

\subsubsection{Training Objective.} According to~\cref{sec:method}, training CanvasMAR involves three objectives: (1) the canvas-masked reconstruction objective for the Spatial MAR head $\mathcal{L}_{\text{MAR}}$; (2) the canvas reconstruction loss $\mathcal{L}_{\text{canvas}}$ in~\cref{eq:canvas_loss}; and (3) the staticness prediction loss $\mathcal{L}_{\text{static}}$ in~\cref{eq:static_loss}. We therefore optimize a weighted objective:
\begin{equation}
\mathcal{L}_{\text{total}} \;=\; \mathcal{L}_{\text{MAR}}\; +\; \lambda_c\, \mathcal{L}_{\text{canvas}}\; +\; \lambda_s\, \mathcal{L}_{\text{static}},
\end{equation}
and set both weights $\lambda_c$ and $\lambda_s$ to $0.1$ in all experiments.

\subsubsection{Tasks and Datasets.} We evaluate CanvasMAR on video prediction tasks, where one or more context frames are given and the goal is to forecast subsequent frames. Our experiments span three datasets of increasing scale and complexity: 1) \textbf{BAIR.} The BAIR dataset~\cite{unterthiner2018towards, ebert2017self} is a toy benchmark containing 43K training videos and 256 evaluation videos of a robot arm pushing objects within a static camera view. 
2) \textbf{UCF-101}. UCF-101~\cite{soomro2012ucf101} is a moderate-scale real-world dataset with over 13{,}000 clips across 101 human action classes. 
3) \textbf{Kinetics-600.}  
The Kinetics-600 dataset~\cite{carreira2018short} is a large-scale collection of diverse real-world videos spanning 600 human action categories. It contains approximately 400K training clips and serves as a challenging benchmark for large-scale video generation.  
Since UCF-101 is typically used for conditional generation without context frames~\cite{zhou2025taming}, which CanvasMAR does not currently support (see clarification in~\cref{sec:conclusion}), we report comparisons only against the baseline method we implemented. For BAIR and Kinetics-600, we provide system-level comparisons with prior art.
See Appendix~\ref{app:details} for more details.
    
\subsection{Component-Level Analysis}

\subsubsection{Effectiveness of the canvas.} 
\begin{figure*}[t]
\centering
\begin{minipage}[c]{0.48\textwidth}
    \centering
    \includegraphics[width=\linewidth]{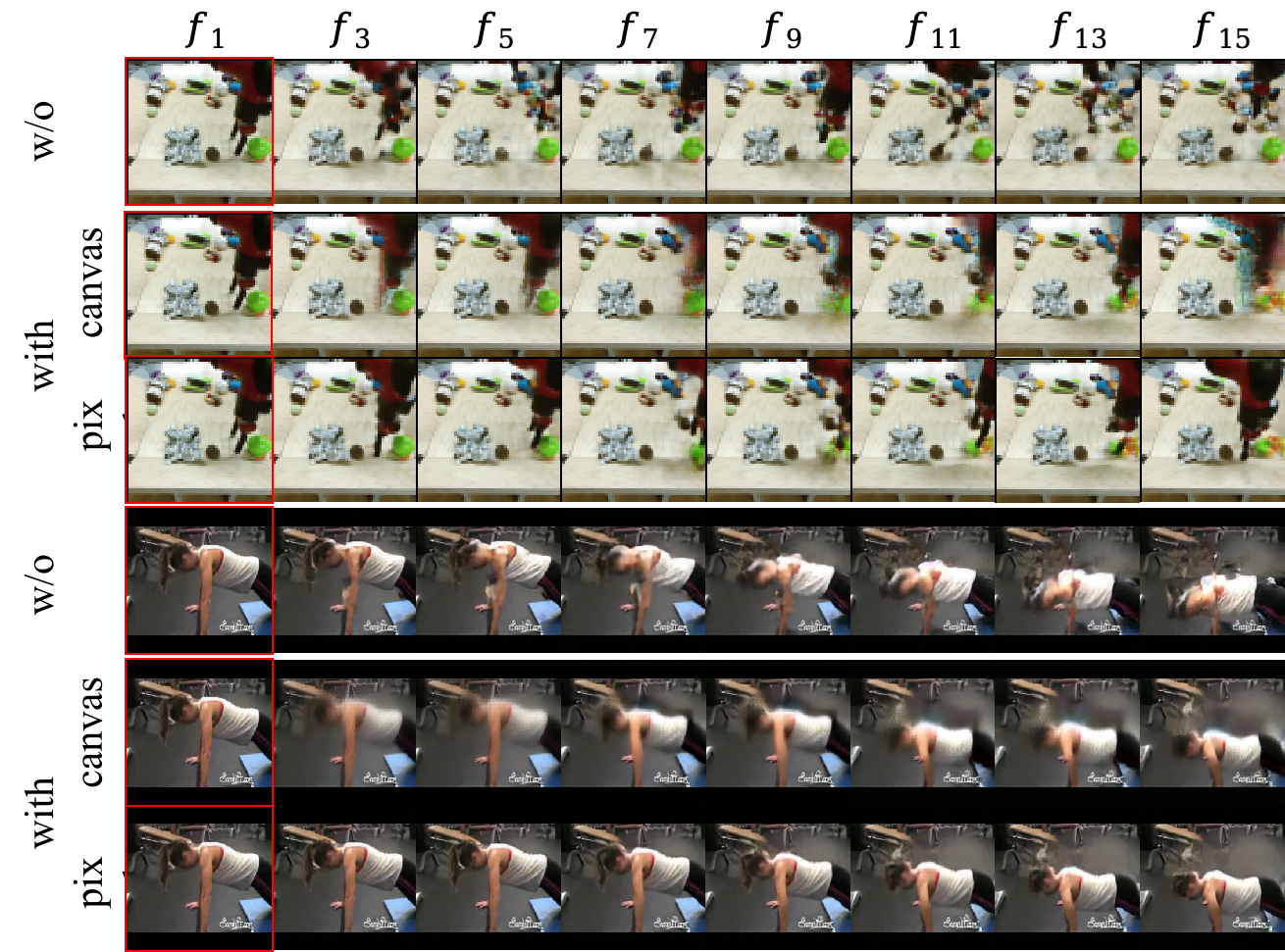}\\[0.25em]
    {\small (a)}
\end{minipage}%
\hfill
\begin{minipage}[c]{0.48\textwidth}
    \centering
    \includegraphics[width=\linewidth]{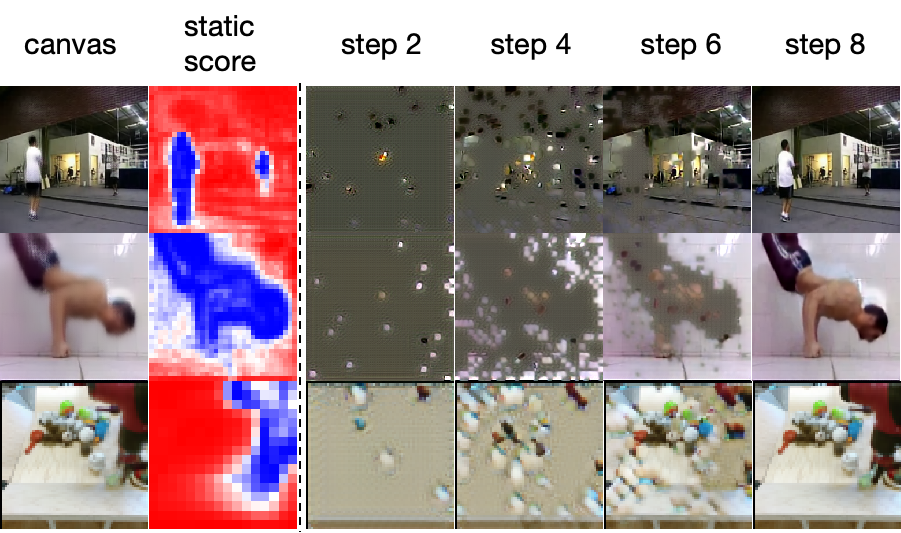}\\[0.25em]
    {\small (b)}
\end{minipage}

\caption{(a) Qualitative examples of video clips generated by CanvasMAR (bottom two rows in each example) and its non-canvas variant (top row). Upper: 2-step generation on BAIR (16 $\times$ 16 latents); Lower: 8-step generation on UCF-101 (32 $\times$ 32 latents). The canvas-based model better preserves global structure and object coherence. (b) Qualitative illustration of the motion-aware staticness prediction and adaptive sampling schedule. Warmer colors denote regions predicted to remain relatively static and are therefore generated earlier, while cooler colors indicate high-motion regions deferred to later autoregressive steps.}
\label{fig:qualitative_combined}
\end{figure*}

\begin{figure}[t]
\centering
\includegraphics[width=0.85\linewidth]{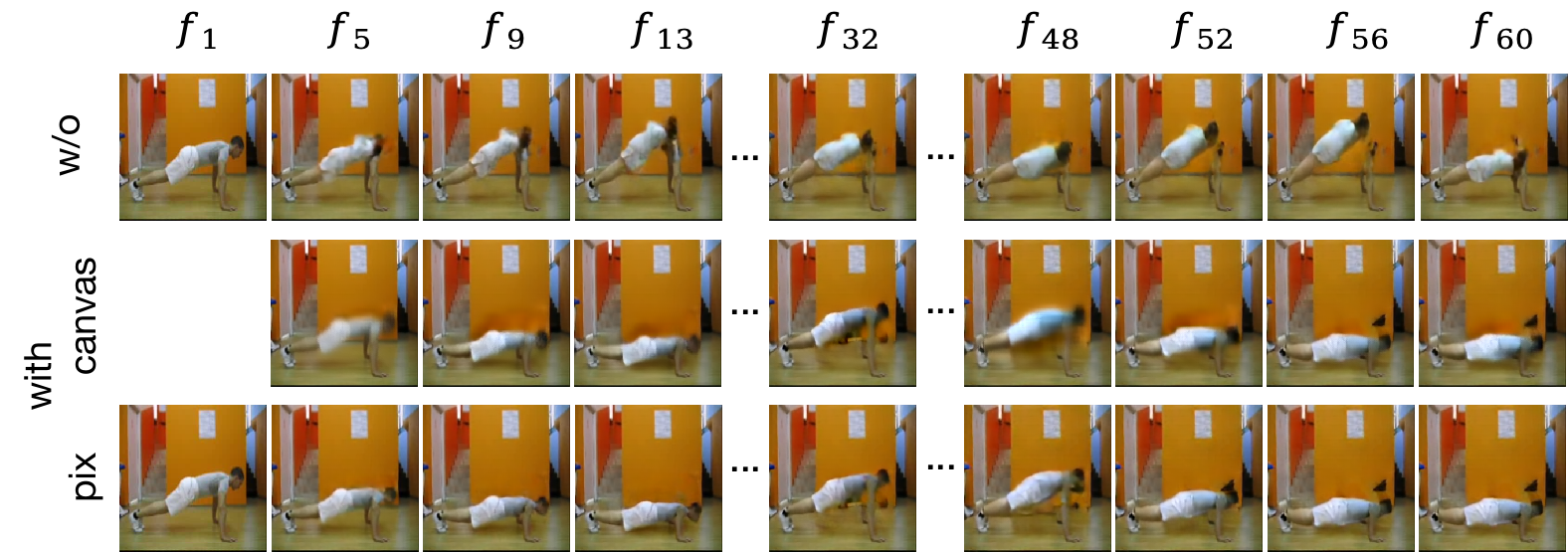}
\caption{Qualitative example of longer video generation generated by CanvasMAR and its non-canvas variant.}
\label{fig:long}
\vspace{-0.5cm}
\end{figure}

Since the canvas module is central to CanvasMAR, we first study its impact. 
To achieve this, we train a model identical to CanvasMAR but without the canvas component \emph{from scratch}, and compare the two under identical experimental settings except for the use of the canvas.  
Results on the BAIR and UCF-101 datasets (\cref{fig:componet_level_analysis} (a) and (b), blue and orange lines) show that the canvas-equipped model consistently outperforms its non-canvas counterpart across all autoregressive step counts. Notably, larger gains appear at small step counts (e.g., 2–4 on the BAIR dataset and 2–8 on the UCF101 dataset), supporting the importance of a global spatial prior in few-step settings.
Qualitative comparisons in~\cref{fig:qualitative_combined} (a) demonstrate that the model without the canvas struggles to maintain global structure and object coherence, whereas the canvas-based model preserves good spatial and temporal consistency in few-step settings. Furthermore, CanvasMAR can generate long videos by autoregressively rolling out frames, and with stable temporal coherence. \cref{fig:long} presents a 60-frame video (compared to the typical 16-frame evaluation) generated by CanvasMAR and its non-canvas counterpart; here, CanvasMAR again demonstrates consistently better generation.

\subsubsection{Effectiveness of adaptive motion-aware sampling.} The adaptive motion-aware sampling strategy enables an easy-to-hard curriculum for next-frame generation. As shown in~\cref{fig:componet_level_analysis} (a) and (b) (orange and green lines), it consistently improves performance in fewer-step settings—the case we focus on for efficient sampling. With more sampling steps, each batch of simultaneously generated tokens becomes smaller, reducing conflicts even without adaptive sampling and therefore narrowing the gap as we expected. Qualitative examples in~\cref{fig:qualitative_combined} (b) visualize the predicted staticness map and its induced ordering, showcasing that the lightweight static head successfully captures motion information. After that, low-motion regions (warm colors) are synthesized first, while high-motion regions (cool colors) are deferred to later steps as expected.

\subsubsection{Effectiveness of compositional CFG.} \cref{fig:componet_level_analysis} (c) summarizes the impact of the spatial and temporal branches of compositional classifier-free guidance. Disabling either branch degrades performance, validating the importance of both guidance for autoregressive video generation.

\subsection{System-Level Comparisons}

We compare CanvasMAR with well-known diffusion-based and autoregressive video models. Note that each method is categorized by its frame-level generation mechanism—either generating full frames or individual image tokens—which more directly determines visual quality than the temporal generation paradigm (generating an entire video as a whole versus rolling out frames sequentially).

\noindent
\begin{minipage}[b]{0.48\textwidth}
\centering
\setlength{\tabcolsep}{4pt}
\captionof{table}{System-level comparison on Kinetics-600. FVD$\downarrow$ at $64\times64$ (methods with $*$ may be evaluated at $128\times128$).}
\label{tab:k600}
\resizebox{\linewidth}{!}{%
    \begin{tabular}{l l r}
    \toprule
    \textbf{Type} & \textbf{Method} & \textbf{FVD}$\downarrow$ \\
    \midrule
    \multirow{2}{*}{GAN} 
    & DVD-GAN-FP~\cite{clark2019adversarial} & 69.1 \\
    & TrIVD-GAN-FP~\cite{luc2020transformation} & 25.74 \\
    \midrule
    \multirow{6}{*}{Diffusion} 
    & VDM~\cite{ho2022video} & 16.2 \\
    & RaMViD~\cite{hoppe2022diffusion} & 16.5 \\
    & Rolling Diffusion~\cite{ruhe2024rollingdiffusionmodels} & 5.2 \\
    & DFoT*~\cite{song2025history} & \textbf{4.3} \\
    \midrule
    \multirow{10}{*}{Autoregressive} 
    & Video Transformer~\cite{weissenborn2019scaling} & 170 \\
    & CogVideo~\cite{hong2022cogvideo} & 109.2 \\
    & Video VQ-VAE~\cite{walker2021predicting} & 64.3 \\
    & CCVS~\cite{le2021ccvs} & 55.0 \\
    & Phenaki*~\cite{villegas2022phenaki} & 36.4 \\
    & Transframer~\cite{nash2022transframer} & 25.4 \\
    & MAGVIT~\cite{yu2023magvit} & 9.9 \\
    & MAGVIT-v2~\cite{yu2023language} & 4.3 \\
    & MAGI~\cite{zhou2025taming} & 11.5 \\
    & \textbf{Ours ($N_{AR}=10$)} & 6.3 \\
    & \textcolor{black!40}{Ours ($N_{AR}=11$)} & \textcolor{black!40}{6.3} \\
    & \textcolor{black!40}{Ours ($N_{AR}=12$)} & \textcolor{black!40}{6.2} \\
    \bottomrule
    \end{tabular}
}
\end{minipage}%
\hfill
\begin{minipage}[b]{0.48\textwidth}
\centering
\setlength{\tabcolsep}{4pt}
\captionof{table}{System-level comparison on BAIR. FVD$\downarrow$ at $64\times64$. DFVD reports debiased FVD~\cite{yu2023magvit}.}
\label{tab:bair}
\resizebox{\linewidth}{!}{%
    \begin{tabular}{l l r r}
    \toprule
    \textbf{Type} & \textbf{Method} & \textbf{FVD}$\downarrow$ & \textbf{DFVD}$\downarrow$ \\
    \midrule
    \multirow{2}{*}{GAN} 
    & DVD-GAN-FP~\cite{clark2019adversarial} & 109.8 & - \\
    & TrIVD-GAN-FP~\cite{luc2020transformation} & 103 & - \\
    \midrule
    \multirow{3}{*}{Diffusion} 
    & MCVD~\cite{voleti2022mcvd} & 90 & - \\
    & RaMViD~\cite{hoppe2022diffusion} & 84 & - \\
    & VDM~\cite{ho2022video} & 66.9 & - \\
    \midrule
    \multirow{8}{*}{AR} 
    & Transframer~\cite{nash2022transframer} & 100 & - \\
    & CCVS~\cite{le2021ccvs} & 99 & - \\
    & Phenaki~\cite{villegas2022phenaki} & 97 & - \\
    & MaskViT~\cite{gupta2022maskvit} & 94 & - \\
    & N\"UWA~\cite{wu2022nuwa} & 87 & - \\
    & MAGVIT-L~\cite{yu2023magvit} & \textbf{62} & 31 \\
    & MAGVIT-B~\cite{yu2023magvit} & 76 & 48 \\
    & \textbf{Ours ($N_{AR}=4$)} & 66.0 & \textbf{26.6} \\
    & \textcolor{black!40}{Ours ($N_{AR}=6$)} & \textcolor{black!40}{65.3} & \textcolor{black!40}{26.5} \\
    & \textcolor{black!40}{Ours ($N_{AR}=8$)} & \textcolor{black!40}{65.2} & \textcolor{black!40}{26.4} \\
    \bottomrule
    \end{tabular}
}
\end{minipage}
\vspace{1em}  

As shown in~\cref{tab:bair}, CanvasMAR achieves the second-best FVD on BAIR among autoregressive models, trailing only MAGVIT. However, under the debiased evaluation, CanvasMAR surpasses MAGVIT, achieving \emph{the best performance}. We argue that the debiased setting is much more convincing due to the wider sample distributions, as detailed in~\cref{app:details}. Notably, MAGVIT is autoregressive only in the spatial domain, but generates all frames simultaneously without enforcing temporal causality. This design reduces the difficulty of video generation but incurs substantially higher latency. In contrast, CanvasMAR generates frames sequentially—making it more prone to error accumulation—yet it delivers better performance.

On the larger and more challenging Kinetics-600 dataset in~\cref{tab:k600}, CanvasMAR trails only MAGVIT-v2~\cite{yu2023language} among autoregressive models, which likewise does not enforce causality and leverages a more advanced autoencoder. Note that MAGI~\cite{zhou2025taming} is a more closely related baseline—both temporally and spatially autoregressive, yet without the canvas mechanism—highlighting the effectiveness of our design.

CanvasMAR also rivals state-of-the-art diffusion models, and with much better efficiency. To illustrate this, we compare inference time with the advanced diffusion model DFoT~\cite{song2025history} in~\cref{fig:inference_and_nextgroup} (a). CanvasMAR is about \textbf{5.7\,$\times$} faster at the largest batch size (GPU fully utilized), highlighting the efficiency advantage of few-step autoregressive sampling. \cref{fig:inference_and_nextgroup} (a) reports latency—the time until the first frame appears—which is often the most relevant metric for interactive applications and one at which autoregressive models typically excel. We further show that even when considering total generation time (see \cref{fig:infer_time_full} in Appendix~\ref{app:details}), CanvasMAR remains about \textbf{2.7\,$\times$} faster than DFoT.

Further increasing the sampling steps of CanvasMAR on both datasets yields only minimal performance gains, indicating that only a few steps are sufficient for CanvasMAR. In contrast, the non-canvas baseline shown in~\cref{fig:componet_level_analysis} fails to surpass the CanvasMAR even with more sampling steps.


\subsection{Next-Group Frame Prediction}

\begin{figure*}[t]
\centering
\begin{minipage}[c]{0.48\textwidth}
    \centering
    \includegraphics[width=\linewidth]{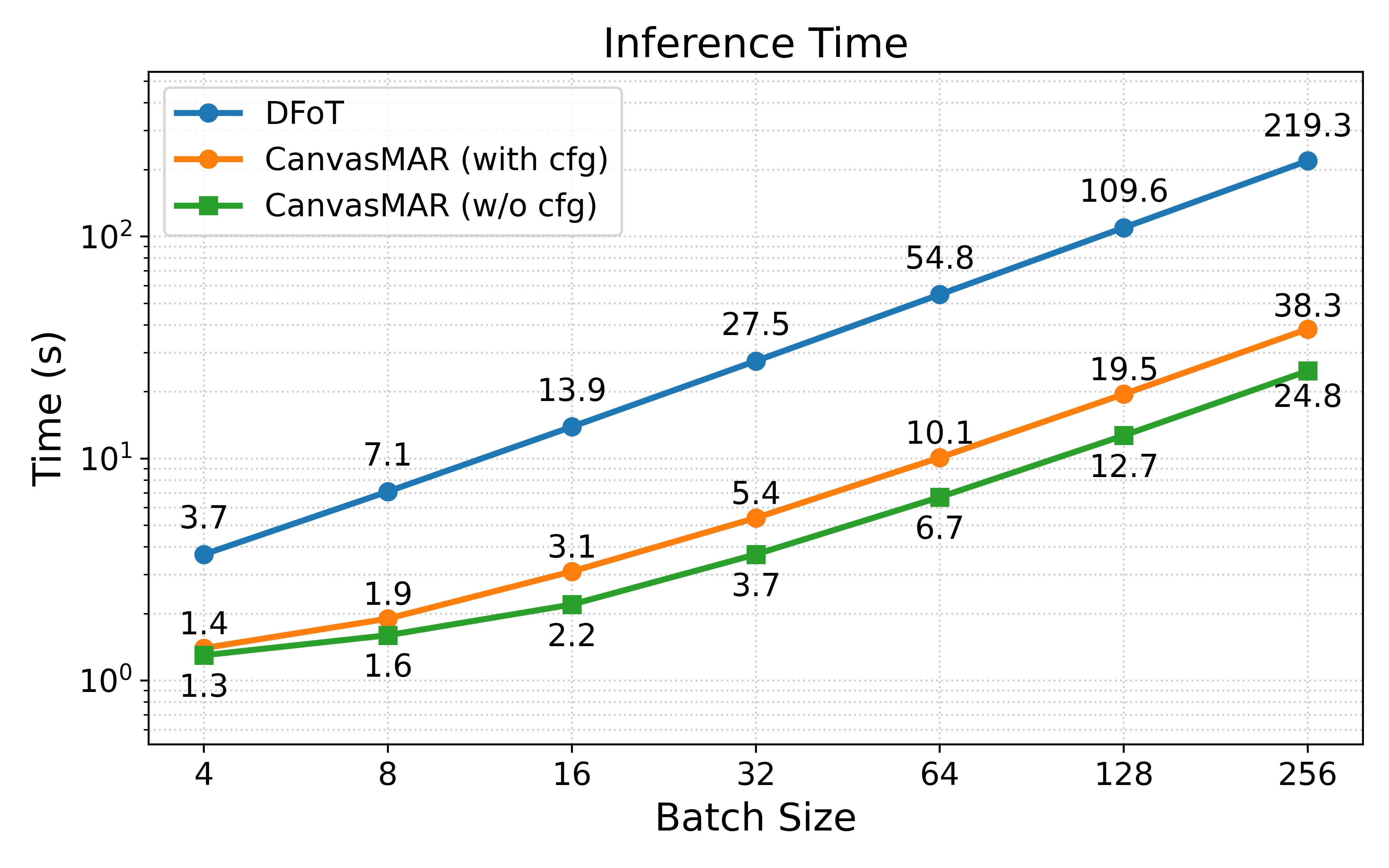}\\[0.25em]
    {\small (a)}
\end{minipage}%
\hfill
\begin{minipage}[c]{0.48\textwidth}
    \centering
    \includegraphics[width=\linewidth]{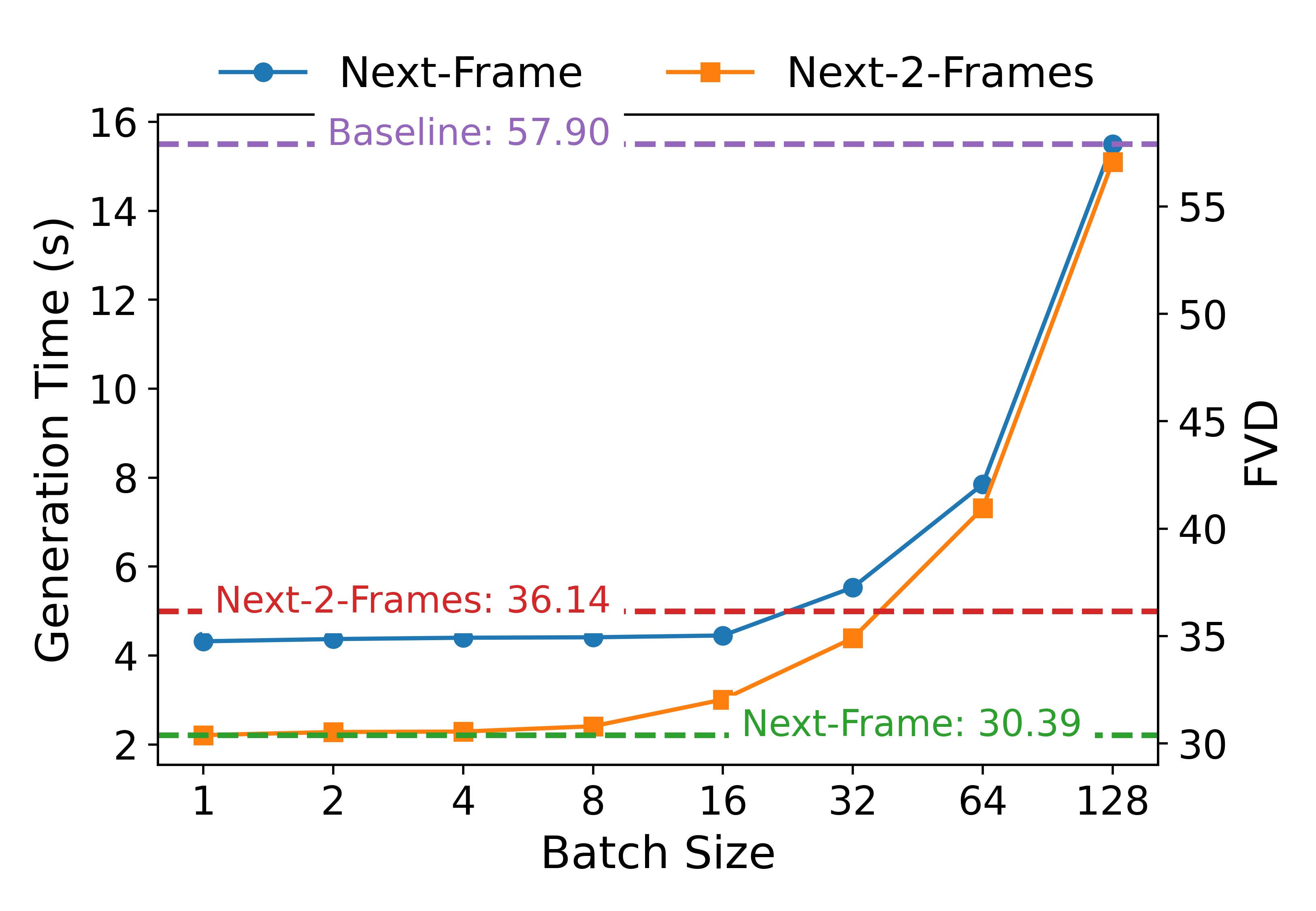}\\[0.25em]
    {\small (b)}
\end{minipage}

\caption{(a) Inference latency comparison of CanvasMAR and DFoT on K600. CanvasMAR achieves much lower latency than DFoT due to the autoregressive paradigm and fewer-step sampling. (b) Performance and speedup effect of next-group prediction models. No classifier-free guidance is used.}
\label{fig:inference_and_nextgroup}
\end{figure*}

In diffusion-based autoregressive models~\cite{yin2025slow, song2025history, huang2025self, liu2025rolling}, predicting multiple upcoming frames (e.g., chunks) jointly is common. For token-based autoregressive models, models typically generate frames one at a time~\cite{deng2024autoregressive}, where next-group prediction often leads to severe performance degradation. However, with the introduction of the canvas, we find next-group frame prediction to be stable, enabling faster generation with limited quality loss. 

Empirically, we fine-tune a next-group frame prediction model (with a group size of 2) from a well-trained next-frame prediction model on BAIR, requiring only about $10\%$ of the original training cost to converge. This indicates that substantial knowledge is transferred between next-frame and next-group settings. See Appendix~\ref{app:next_group_prediction} for more details of the model architecture.

As shown in~\cref{fig:inference_and_nextgroup} (b), the next-2-frame model exhibits substantial speedups at smaller batch sizes compared to the next-frame model and converges to similar throughput at higher batch sizes. When the batch size becomes large, both models fully utilize the available GPU resources and, thanks to KV caching, reach comparable parallel efficiency. In practical scenarios, however, users typically generate only a few videos and expect low latency, where the next-2-frame prediction model is particularly advantageous. Furthermore, we train a next-2-frame prediction model {without} the canvas (purple curve), and our canvas-based model performs significantly better when transferred to next-group prediction because the canvas provides strong structural priors for multiple future frames.

\section{Conclusion and Limitations}
\label{sec:conclusion}

\subsubsection{Conclusions.} We presented CanvasMAR, a masked autoregressive video prediction framework that improves both quality and efficiency through a canvas-based conditioning mechanism. By generating a blurred global estimate of the next frame before spatial token generation, CanvasMAR bridges the gap between fast temporal and slow spatial autoregression, reducing the generation step in conventional MAR models and improving fidelity. Experiments on BAIR and Kinetics-600 confirm that CanvasMAR achieves competitive or superior FVD scores with fewer autoregressive steps, outperforming most prior autoregressive baselines.

\subsubsection{Limitations.} Despite its strong performance, CanvasMAR has several limitations. First, CanvasMAR is primarily evaluated on video prediction tasks, which assume the availability of previous frames. This focus stems from the core contribution of the paper—the canvas mechanism—which relies on preceding frames. However, this is not an inherent limitation of CanvasMAR: it can also generate videos from scratch, as the first frame can always be generated without the canvas mechanism (similar to NOVA~\cite{li2024autoregressive}). We did not evaluate this scenario as it lies outside the main contribution of our work. Second, when predicting videos with substantial motion, CanvasMAR can produce severely distorted results, as shown in~\cref{fig:fail} in Appendix~\cref{app:details}. One reason is that the initial canvas can be overly blurred for such sequences, which misleads the Spatial MAR and leaves it unable to fully correct these artifacts. Scaling up the model may improve its self-correction ability, especially considering that our largest Spatial MAR has only half the layers of NOVA~\cite{deng2024autoregressive}. Future work will also explore scaling CanvasMAR to text-conditioned and multimodal settings.

%
%
\bibliographystyle{splncs04}
\bibliography{main}

\clearpage
\newpage
\appendix
\section{Experiment Details}
\label{app:details}

\subsubsection{Training and Model Details.} All models are trained on NVIDIA A100 GPUs. Comprehensive model configurations and hyperparameters are listed in~\cref{tab:model-configs}.

\subsubsection{Sampling Protocols}
As noted by Skorokhodov \etal~\cite{skorokhodov2022stylegan}, the sampling procedure can significantly influence FVD scores. To ensure fair comparisons, we follow the sampling strategies used in prior work:

\begin{itemize}
\item \textbf{BAIR}: For each of the 256 test videos, we use the first frame as the conditioning input and the subsequent 16-frame segment as the real distribution. We then draw $K$ synthetic 16-frame clips per conditioning frame, resulting in $256 \times K$ generated clips in the fake distribution.

Following Yu \etal~\cite{yu2023magvit}, we also report the \textit{debiased FVD} by randomly sampling one conditioning frame per test video and redefining the real distribution accordingly. Repeating this procedure 100 times per video again yields $256 \times K$ generated clips.

Unless otherwise specified, we set $K=20$ and report only the debiased FVD. For system-level comparisons, we align with prior work by using $K=100$ and reporting both standard and debiased FVD scores.

\item \textbf{UCF-101}: We randomly sample 5K test videos and, for each, select a random 16-frame clip to define the real distribution, using the preceding 1 frame as a condition to generate one 16-frame clip for the fake distribution.

\item \textbf{Kinetics-600}: We randomly sample $K$ test videos and, for each, select a random 16-frame clip to define the real distribution, using the preceding 5 frames as conditions to generate one 16-frame clip for the fake distribution. For the system-level study, $K=50000$, which aligns with previous works. For ablation studies, we set $K=10000$.
\end{itemize}

\subsubsection{Inference Time} 
\begin{figure}[t]
\centering
\includegraphics[width=0.75\linewidth]{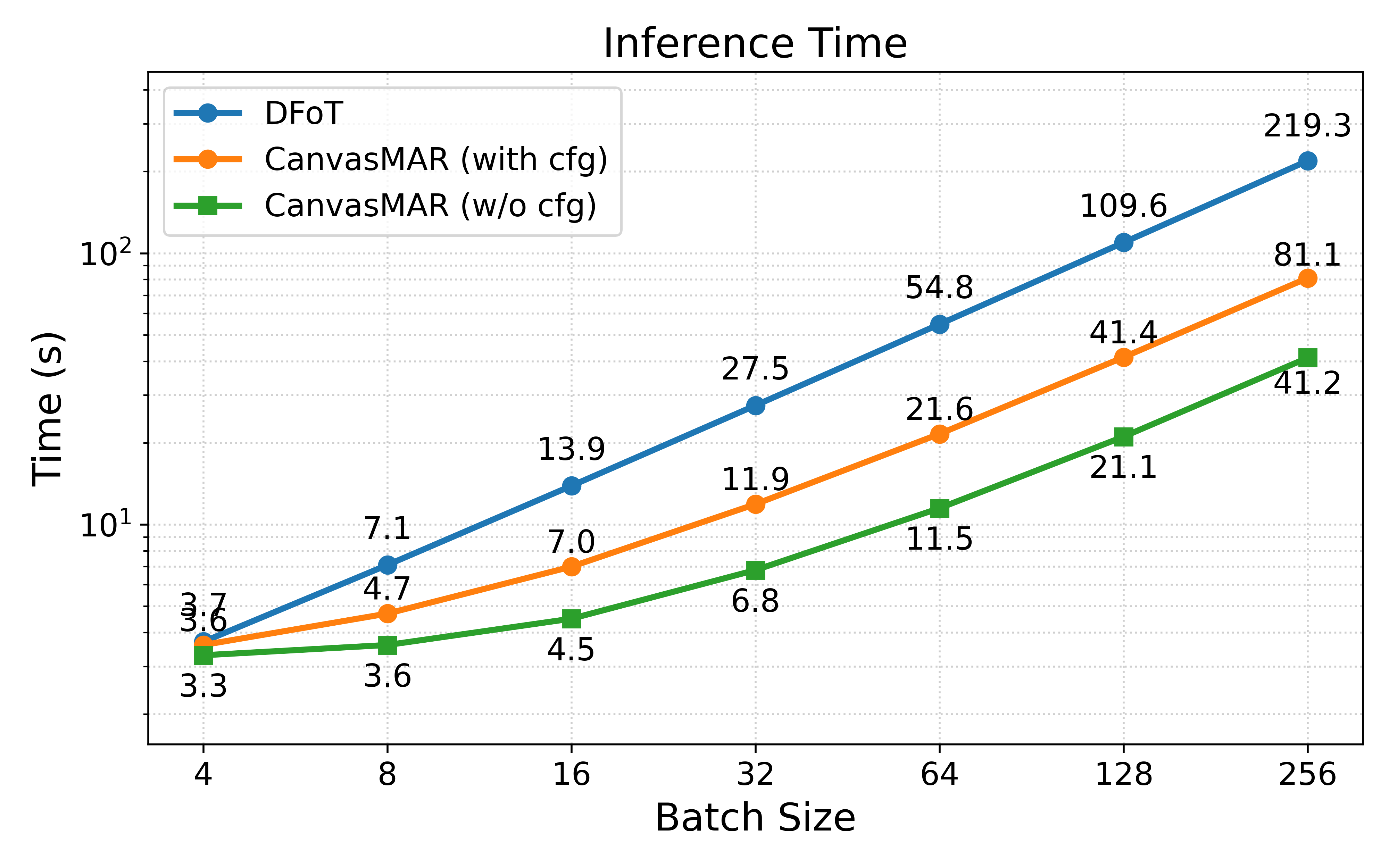}
\caption{Inference time comparison of CanvasMAR and DFoT on Kinetics-600. CanvasMAR achieves much lower generation time than DFoT due to the fewer-step sampling.}
\label{fig:infer_time_full}
\end{figure}

For completeness, we report the full generation time comparison in~\cref{fig:infer_time_full}.

\subsubsection{Failure Case of CanvasMAR} 
\begin{figure}[ht]
\centering
\includegraphics[width=\linewidth]{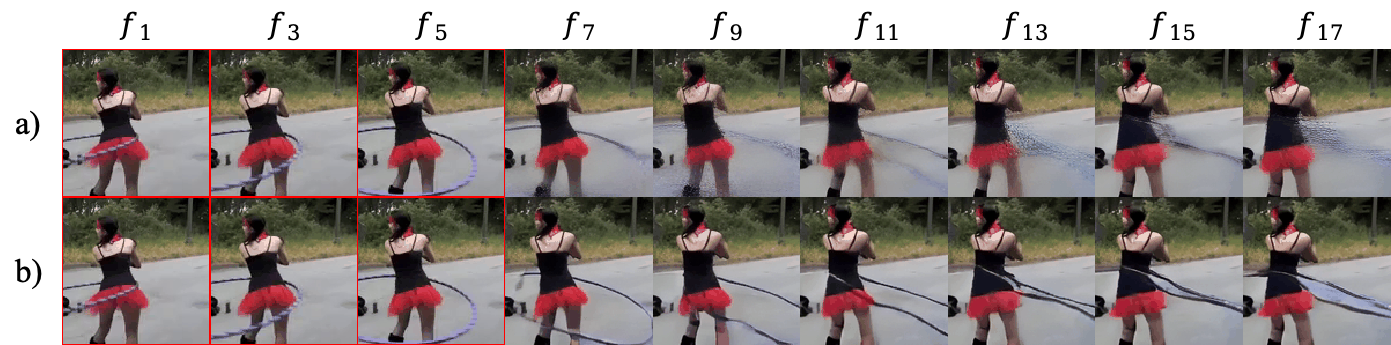}
\caption{Failure case of CanvasMAR for videos exhibiting substantial motion. a) The predicted canvas. b) The generated frame.}
\label{fig:fail}
\end{figure}

CanvasMAR show sub-optimal performance on highly dynamic video predictions. See~\cref{fig:fail}.

\begin{table*}[t]
\centering
\resizebox{\linewidth}{!}{
\begin{tabular}{lccc}
\toprule
\textbf{Component} & \textbf{BAIR} & \textbf{UCF-101} & \textbf{Kinetics-600} \\
\midrule
\multicolumn{4}{@{}l}{\textbf{Architecture}} \\
Temporal ViT (dim / layers / patch) & 768 / 8 / 2 & 1024 / 16 / 2 & 1024 / 16 / 2 \\
Canvas ViT (dim / layers / patch)   & 768 / 4 / 1 & 1024 / 8 / 1 & 1024 / 8 / 1 \\
Spatial MAR (dim / layers / depth / patch) & 768 / 8 / 2 / 1 & 1024 / 16 / 4 / 1 & 1024 / 16 / 4 / 1 \\
Additional Architecture & Expert Layernorm~\cite{yang2024cogvideox} & Expert Layernorm~\cite{yang2024cogvideox} & MM-DiT~\cite{esser2024scaling} \\
Flow decoder (dim / layers) & 1280 / 3 & 1280 / 3 & 1280 / 3 \\
VAE model & Causal 3D CNN$^*$ & OmniTokenizer~\cite{wang2024omnitokenizer} & Non-causal 3D CNN~\cite{song2025history} \\
MAR autoregressive steps$^{\#}$ & 4 & 8 & 10 \\
Total parameters & 211M & 733M & 733M \\
\midrule
\multicolumn{4}{@{}l}{\textbf{Training}} \\
Batch size & 192 & 64 & 192 \\
Epochs & 600 & 800 & 270 \\
Resolution & $17 \times 64 \times 64$ & $17 \times 256 \times 256$ & $17 \times 128 \times 128$ \\
Latent compression ratio & $\{1, 2\} \times 4 \times 4$ & $\{1, 4\} \times 8 \times 8$ & $\{1, 4\} \times 8 \times 8$ \\
Latent resolution & $9 \times 16 \times 16$ & $5 \times 32 \times 32$ & $5 \times 16 \times 16$ \\
Noise augmentation ($r$) & $[0.2, 0.6]$ & $[0.2, 0.6]$ & $[0.2, 0.6]$ \\
Noise augmentation ($r'$) & $[0.0, 0.3]$ & $[0.0, 0.3]$ & $[0.0, 0.3]$ \\
\midrule
\multicolumn{4}{@{}l}{\textbf{Sampling}} \\
Noise augmentation ($r$) & $0.3$ & $0.3$ & $0.4$ \\
Noise augmentation ($r'$) & $0.2$ & $0.2$ & $0.2$ \\
Confidence temperature & 6.5 & 6.5 & 6.5 \\
Flow-head steps & 30 & 30 & 30 \\
Canvas CFG & 1.5 & 1.5 & 2.25 \\
Temporal CFG & 1.2 & 1.2 & 1.1 \\
Evaluation Resolution & $16 \times 64 \times 64$  & $16 \times 256 \times 256$  & $16 \times 64 \times 64$  \\
\bottomrule
\end{tabular}
}
\caption{
Model configurations for BAIR, UCF-101, and Kinetics-600. $^*$ Trained by ourselves with codebase of Song \etal~\cite{song2025history}. $^{\#}$ Otherwise specified, we use this step number for sampling.
}
\label{tab:model-configs}
\end{table*}

\begin{figure}[h]
\centering
\includegraphics[width=0.7\linewidth]{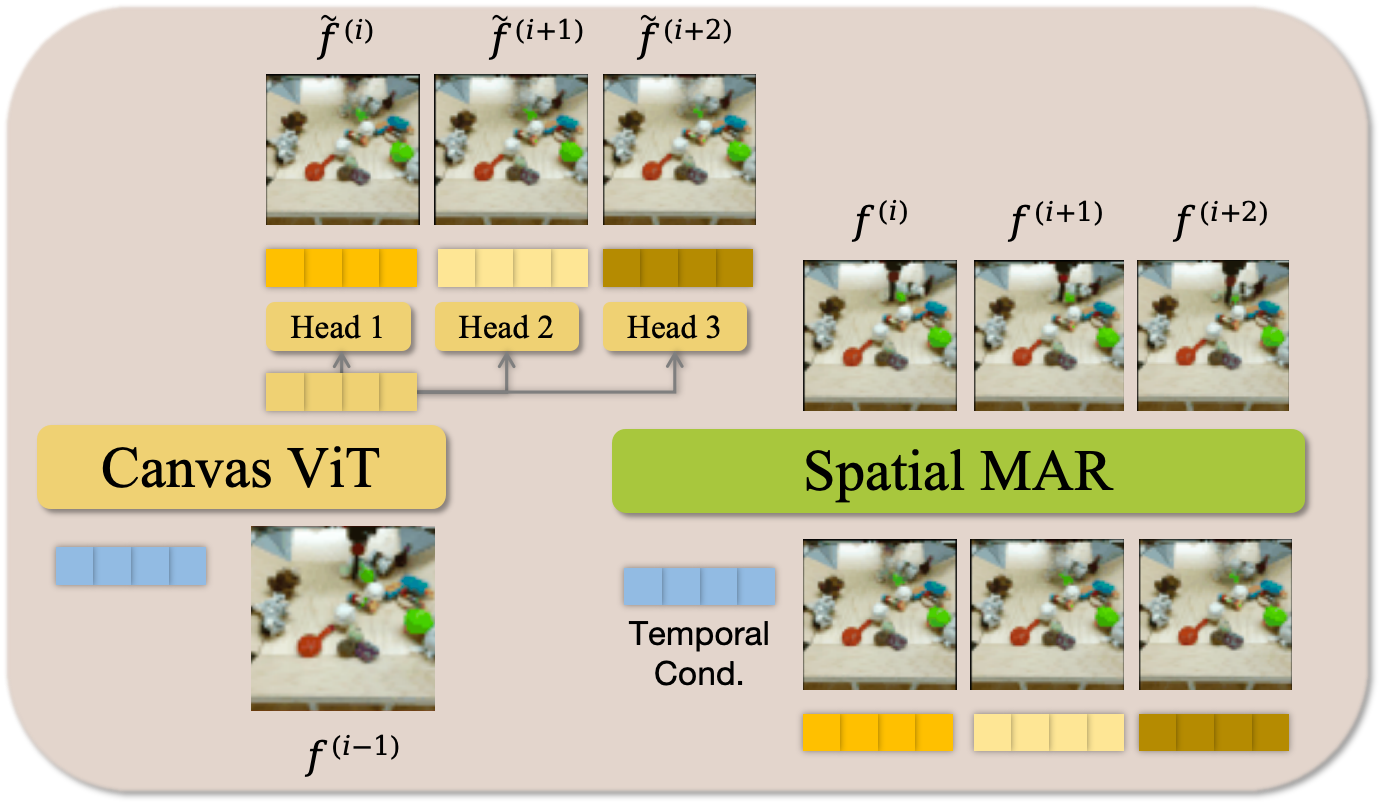}
\caption{Next-group frame prediction.}
\label{fig:next_group}
\end{figure}

\subsubsection{Canvas Augmentation} The canvas is predicted in a single pass, unlike iterative generative models that refine outputs across multiple steps. This design introduces two primary sources of error.  
First, inaccuracies in the previously generated frame can mislead the Canvas ViT, compounding errors in the next canvas prediction.  
Second, even with an accurate previous frame, the one-pass Canvas ViT may produce overly blurred or imprecise canvases, which can confuse the Spatial MAR and amplify artifacts.

To mitigate these effects, we additionally adopt a noise-based augmentation strategy inspired by cascaded generation~\cite{zhang2025flashvideo, ho2022cascaded}.  
When feeding the most recent frame $f^{(i-1)}$ into the Canvas ViT, we add Gaussian noise:
\begin{equation}
    f^{(i-1)}_{\text{aug}} = f^{(i-1)} \cdot (1 - r) + \epsilon \cdot r,
\end{equation}
where $\epsilon$ denotes Gaussian noise and $r$ is an interpolation coefficient.  
Similarly, we perturb the canvas embedding:
\begin{equation}
    z_{c,\text{aug}}^{(i)} = z_c^{(i)} \cdot (1 - r') + \epsilon' \cdot r',
\end{equation}
where $\epsilon'$ and $r'$ denote another Gaussian noise term and interpolation coefficient.  
These augmentations simulate instability in previously generated frames and canvases, encouraging both the Canvas ViT and the Spatial MAR to remain robust under imperfect conditions.  
During training, we sample $r$ and $r'$ uniformly from a predefined interval; during inference, we fix both to a constant. See~\cref{tab:model-configs} for the specific values used in our experiments.

\section{Next-Group Frame Prediction}
\label{app:next_group_prediction}

For the next-$K$-frame prediction model, we slightly modify the architecture by appending $K$ \emph{canvas heads}, each implemented as a two-layer MLP, to the output of the Canvas ViT, as shown in~\cref{fig:next_group}. We also add lightweight temporal attention layers among spatial layers every four layers, following the decomposed attention design of Gupta \etal~\cite{gupta2024photorealistic}, to ensure temporal consistency. These canvas heads provide the spatial conditioning canvases for generating the next $K$ frames within the Spatial MAR.

\end{document}